# Self-adjusting optimization algorithm for solving the set-union knapsack problem

*Congcong Wu[1,2], Xiangyun Gao[1*], Xueyong Liu[3], Bowen Sun[4]*
1. School of Economics and Management, China University of Geosciences, Beijing 100083, China
2. College of Information Engineering, Hebei GEO University, Shijiazhuang Hebei 050031, China
3. School of Management and Engineering, Capital University of Economics and Business, Beijing, 100070, China
4. School of Business, Henan Normal University, Xinxiang Henan 453007, China
*Corresponding author: Xiangyun Gao, E-mail: gxy5669777@126.com*

**Abstract:** The set-union knapsack problem (SUKP) is a constrained composed optimization problem. It is more difficulty for solving because values and weights depend on items and elements respectively. In this paper, we present two self-adjusting optimization algorithms for approximating SUKP from items and elements perspective respectively. By analyzing the dynamic characters in the SUKP, we design two types of self-adjusting repair and optimization operators that are based on the different loading process. We use the novel teaching-learning-based optimization algorithm (TLBO) to design a general discrete framework (DTLBO) suitable for these two types of operators. In addition, we introduce elite opposite search and natural selection mechanism into DTLBO to furtherly improve the performance of the algorithm from the perspective of population. Finally, we performed experimental comparisons on benchmark sets to verify the effectiveness of the proposed algorithm. The experimental results show that the item-based self-adjusting optimization algorithm I-DTLBO is outstanding, and the algorithm is superior to the other swarm intelligence methods for solving SUKP. I-DTLBO algorithm reaches the upper boundary of the current swarm intelligence algorithms for solving SUKP in 10 instances, and gotten new upper boundary in 15 instances. The algorithm E-DTLBO based on element loading only perform slightly better on small and middle data sets, but worse on large-scale instances. It shows that element-based design is not suitable for solving SUKP.

adjusting repair and optimization operator · teaching-learning-based optimization algorithm

# 1. Introduction

The set union knapsack problem (SUKP) (Arulselvan and Ashwin 2014; Goldschmidt et al. 1994) is a special type of knapsack problem and is more complicated than the classical 0-1 knapsack problem. There is an element set $\mathbf{E} = \{e_1, e_2, ..., e_n\}$ and an item set $\mathbf{S} = \{\mathbf{U}_1, \mathbf{U}_2, ..., \mathbf{U}_m\}$ in SUKP. Each item $\mathbf{U}_i$ ($i = 1, 2, ..., m$) in $\mathbf{S}$ is a subset of $\mathbf{E}$, where $\mathbf{U}_i \subseteq \mathbf{E}$. Each item $\mathbf{U}_i$ has one nonnegative value $p_i$ and each element $e_j$ has one nonnegative weight $w_j$ ($j = 1, 2, ..., n$). The objective is to select items from $\mathbf{S}$ to load into the knapsack such that the sum of the weights of the elements in the knapsack does not exceed the capacity $C$, while maximizing the sum of the values of the items. It can be expressed as follows (Arulselvan and Ashwin 2014; Goldschmidt et al. 1994):

$$Maximize \ P(\mathbf{A}) = \sum_{\mathbf{U}_i \in \mathbf{A}} p_i \qquad (1)$$

$$Subject \ to \ \ W(\mathbf{A}) = \sum_{e_j \in \bigcup_{\mathbf{U}_i \in \mathbf{A}} \mathbf{U}_i} w_j \leq C, \ \mathbf{A} \subseteq \mathbf{S} \qquad (2)$$

where $\mathbf{A}$ is the set of items in the knapsack.

SUKP has many applications, such as allocation in databases (Goldschmidt et al. 1994; Navathe et al. 1984), budget scenario problems (Jagiello J 2017), flexible manufacturing systems (Crama 1997; Goldschmidt et al. 1994; Hirabayashi et al. 1984), caching system for animation (Lister et al. 2010), and problem of selecting cloud services to migrate (Diallo et al. 2016). SUKP was proposed by Goldschmidt Olivier et al. in 1994 and they regard SUKP as a generalization of the 0-1 knapsack problem (Goldschmidt et al. 1994). A dynamic programming algorithm was proposed for solving SUKP in the paper. However, the dynamic programing algorithm runs in polynomial time only for special cases (Goldschmidt et al. 1994). To the best of our knowledge, this is the only study that uses dynamic programming to solve SUKP. SUKP has difficultly satisfying the principle of optimality because the optimal decision for each of the remaining stages in SUKP depends on previous decisions. After 20 years, Arulselvan proposed an approximation algorithm (A-SUKP) (Arulselvan and Ashwin 2014). It is based on a greedy strategy and its approximation rate is 1-$e^{-1/d}$, where $d$ ($d \geq 2$) is the upper bound of the occurrence of all the elements. However, if $d$ is large, the approximation solution of A-SUKP is unsatisfactory (Arulselvan and Ashwin 2014). In recent years, scholars have begun to apply the swarm intelligence algorithm to solve SUKP and have obtained satisfactory results. He et al. (He et al. 2018a) used binary artificial bee colony algorithm (BABC) to solve SUKP, and the computing speed and accuracy were both higher than those of Ashwin's approximation algorithm. Afterwards, they proposed a new algorithm, namely, the group-theory-based optimization algorithm (GTOS), and applied it to solve SUKP (He et al. 2018b). The new algorithm outperformed previous algorithms in solving SUKP. In 2018, Fehmi B. Ozsoydan and Adil Baykasoglu proposed a GA and PSO hybrid algorithm, namely, gPSO, for solving SUKP (Ozsoydan and Adil 2018). The gPSO solved SUKP with higher accuracy. However, the stability of the solution is not satisfactory. In 2019, Feng et al. tested twelve discrete moth search algorithms for addressing SUKP (Feng et al. 2018). Liu and He applied estimation of distribution to solve SUKP (Liu and He 2019). In our previous work (Wu and He 2020), we designed a hybrid Jaya algorithm for solving SUKP, which increased accuracy and stability of the solution. In addition to the swarm intelligence algorithms, Wei and Hao presented an iterated two-phase local search algorithm for SUKP (Wei and Hao 2019), which discovered some better results. According to previous research, the swarm intelligence algorithm is the optimal and most practical choice for solving SUKP. But, the accuracy and stability of the solution for SUKP is still to be improved. This is also the goal of our research in this work.

The SUKP is a constraint composition optimization problem. A large number of infeasible solutions will be generated in the process of searching using the swarm intelligence algorithm. Hence, the repair of infeasible solutions in evolution is indispensable and it plays a key role in improving the accuracy of the solution. Previous studies used mostly the S-GROA (Feng et al. 2018; He et al. 2018a; He et al. 2018b; Ozsoydan and Adil 2018) or the modified S-GROA (Wu and He 2020) to repair and optimize the solutions. However, the dynamic correlation character in SUKP was not considered in these repair and optimization strategies. The loading of the previous item affects the outside items because the previous item may contain elements belong to items that are outside the knapsack. To this end, we propose the self-adjusting repair and optimization strategy of individuals based on dynamic character of SUKP. The strategy substantially improves the accuracy of solution by optimization of each individual. In addition, the previous research considers the loading process from the perspective of the item. In SUKP, the item has value and each element has weight. Because the item consists of elements, the element is the smallest unit in SUKP. Loading from the perspective of the element is expected to yield a more accurate solution. In this work, we explore a new method based on element loading.

Swarm intelligence algorithms, due to their inherent global searching abilities, have been widely used to solve complex optimization problems and have yielded remarkable results. The teaching-learning-based optimization algorithm (TLBO), a novel swarm intelligence algorithm, was proposed by Rao et al. (Rao et al. 2011a; Rao et al. 2011b). TLBO has attracted the attention of many scholars and has been applied to various optimization problems (Kamel et al. 2018; Li et al. 2020; Rao and Patel 2013; Rao and Rai 2016) due to its simple structure and efficient performance. We design a general discrete TLBO architecture to solve SUKP. In order to further improve accuracy and stability of the solution, the original TLBO is improved. We introduce an elite opposite search strategy into TLBO for helping the algorithm to jump out of local trap. In addition, the natural selection mechanism is employed to increase the diversity of the population and enhance the global search capability.

The remainder of the paper is organized as follows: In Section 2, we analysis the dynamic character in SUKP and propose two self-adjusting repair and optimization operators. In Section 3, we present a universal improved discrete TLBO framework for two loading methods to solve SUKP. In Section 4, the experimental results are compared with the state-of-the-art swarm intelligence algorithms for solving SUKP. Finally, the conclusions of this work are presented and future work is discussed in Section 5.

# 2 Two self-adjusting repair and optimization operators

In previous studies on SUKP, no one considered the changes in the relative characteristics of item and element during the loading process. Capturing this dynamic feature and taking necessary measures will improve the accuracy of solving SUKP. In addition, scholars typically design individuals from the items when using a swarm intelligence algorithm to solve SUKP. In this work, we not only study the method based on item as the loading unit, but also try to use the element as the loading unit.

## 2.1 Analysis of the dynamic characteristic of SUKP

To be intuitive, we can describe the relationship between items and elements of SUKP in the form of a graph (As shown in Figure 1). The Figure 1(a) is initial state in a SUKP instance. The edges connect the items and the elements when they have containing relationship. For example, U1 is composed of elements e1, e3 and e4. On the other hand, e1 belongs to U1 and U2 at the same time. From the perspective of loading items, when we choose to load U1 into the knapsack, it is equivalent to selecting elements e1, e3 and e4. Because the two elements, e1 and e4, are included in U2 too, only one e5 is outside for U2 when U1 is loaded. At this time, U2 only needs to consider the weight of e5. This means that although some items are not loaded into the knapsack, a part of them have already entered the knapsack (see the change from Figure 1(a) to Figure 1(b)). It can be seen that the item in the knapsack will affect the relative weight of other outside items. This dynamic correlation in SUKP has never been considered by previous researchers. But it is vital to individual repairing and optimization when we use the swarm intelligence algorithm to solve SUKP. In this work, we design the self-adjusting repair and optimization operator according to the dynamic correlation of SUKP. So as to select the objects to be loaded more reasonably, thereby improving the individual's optimization ability.

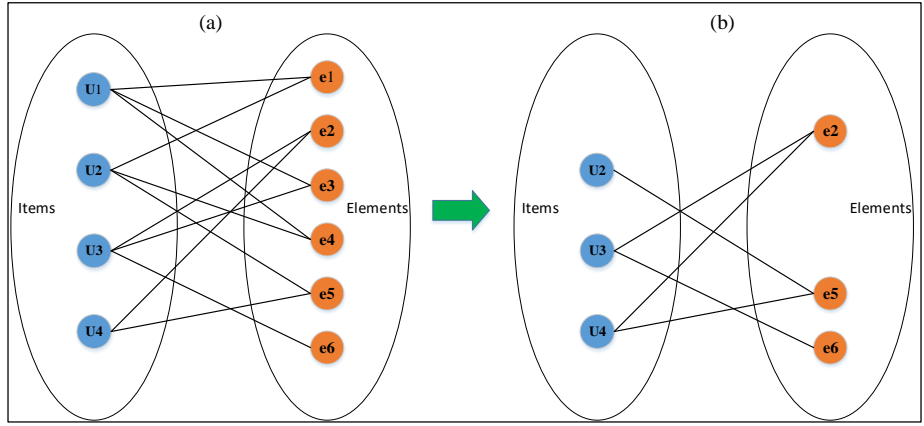

**Fig. 1** The dynamic characteristic of SUKP: (a) The state of SUKP when no item is loaded to knapsack, (b) The state of SUKP when the item **U**1 is loaded to knapsack.

## 2.2 Item-based loading and the self-adjusting repair and optimization operator

In the item-based loading, each solution **Y** = [$y_1$, $y_2$, ..., $y_m$], where $y_j \in \{0, 1\}$, specifies whether an item is loaded or not: $y_j$ =1 indicates that the $j$th item in **S** is selected for loading; otherwise, it is not loaded. If **Y** satisfies Eq. (2), it is a feasible solution; otherwise, it is an infeasible solution. We call the solution that is generated in evolution a candidate solution because the feasibility of the solution is not known.

When using swarm intelligence algorithm to solve SUKP, the method often used to repair infeasible solutions and optimize feasible solutions is greedy. Generally, the items are sorted in descending order according to the value density of the items, and then greedy corrections are made to the individual based on the characteristics of the individual and this descending sequence (Feng et al. 2018; He et al. 2018a; He et al. 2018b; Liu and He 2019; Ozsoydan and Adil 2018; Wu and He 2020). However, these methods ignore dynamic correlation of items in SUKP. They consider value density of the item to be constant. But in fact, as we analyze in section 2.1 that some elements of items outside knapsack will be loaded into the knapsack along with other items, which causes the value density of the items outside to rise. So, we designed the item-based self-adjusting repair and optimization operator (ISRO) based on this dynamic characteristic of SUKP. ISRO repairs the infeasible solutions and optimizes the feasible solutions according to their relative value densities.

**Define**:

**Frequency of an element (*FE*)**: the total number of times that the element appears in items.

**Unit weight of an element (*UWE*)**: the ratio of the weight of the element to its *FE*.

**Absolute value density of an item** (*AVDI*): The value of the item divided by the total *FE* of the elements in the item.

$$AVDI = p_i \bigg/ \sum_{e_j \in \mathbf{U}_i} w'_j \tag{3}$$

**Relative weight of an item (*RWI)***: The sum of the *UWEs* of the elements that belong to the item and are not in the knapsack.

**Relative value density of an item (*RVDI*)**: The value of the item divided by the *RWI* of the item.

$$RVDI = p_i \bigg/ \sum_{e_j \in \mathbf{U}_i, e_j \notin \bigcup_{\mathbf{U}_k \in \mathbf{A}} \mathbf{U}_k} w'_j \tag{4}$$

In Eq. (3) and (4), $p_i$ is the value of item $\mathbf{U}_i$, $w'_j$ is the *UWE* of element $e_j$, and $\mathbf{A}$ is the set of items that are in the knapsack. In the previous swarm intelligence algorithm for solving SUKP, greedy repair and optimization strategies were conducted according to the *AVDI*. In ISRO, greedy loading is conducted according to the *RVI*. ISRO is described as follows.

---
**Algorithm 1**: ISRO
**Input:** The candidate solution $\mathbf{Y} = [y_1, y_2, ..., y_m]$, $y_j \in \{0, 1\}$.
**Output:** The optimized feasible solution $\mathbf{Y} = [y_1, y_2, ..., y_m]$, $y_j \in \{0, 1\}$; the total value of the items in the knapsack, namely, the value of the objective function.

---
1). Let all items with a value of 1 in **Y** form set **S**1 and let all items with a value of 0 in **Y** form set **S**2. Let **Y** be a zero vector.
2). **while** ( $\mathbf{S}1 \neq \Phi$ )
3).     Calculate the *RVDI* of each item in **S**1
4).     Identify the item $\mathbf{U}_k$ with the highest *RVI* in **S**1
5).     **if** $(W(\mathbf{A} \cup \mathbf{U}_k) \leq C)$    **then**    $\mathbf{A} = \mathbf{A} \cup \mathbf{U}_k$, $y_k = 1$

| | |
|---|---|
| 6). | S1=S1- $U_k$ |
| 7). | **end while** |
| 8). | **while** ( S2 ≠ Φ ) |
| 9). | Calculate the *RVDI* of each item in S2 |
| 10). | Identify the item $U_k$ with the highest of *RVI* in S2 |
| 11). | **if** (W(A ∪ $U_k$) ≤ C)   **then**   A= A ∪ $U_k$, $y_k$=1 |
| 12). | S2=S2 - $U_k$ |
| 13). *end while* | |
| 14). *return* (Y, P(A)) | |

ISRO selects **Y** as a feasible solution by conducting step 2 to step 7 and optimizes **Y** to obtain a higher quality solution by executing step 8 to step 13.

## 2.3 Element-based loading and its self-adjusting repair and optimization operator

In SUKP, the item is the basic carrier of value and the element is the basic carrier of weight. Items are composed of elements. Hence, elements are smaller units. We guess that a more granular loading method will result in a more accurate solution. Therefore, we designed a scheme for loading units with elements. In the element-based load design, each solution **Y** = [$y_1$, $y_2$, ..., $y_n$], where $y_j \in \{0, 1\}$, specifies whether each element will be loaded or not: $y_j$=1 indicates that the *j*th element from **E** will be selected for loading; otherwise, it will not be loaded. A mathematical representation of SUKP from the perspective of the element is as follows:

$$Maximize\ P(\mathbf{Y}) = \sum_{i=1}^{n}(\prod_{\forall e_j \in U_i} y_j)p_i \qquad (5)$$

$$Subject\ to\ \begin{cases} \sum_{j=1}^{m} y_j w_j \leq C \\ \forall y_j = 1 \rightarrow \exists U_i \in \mathbf{A}, e_j \in U_i \\ y_j \in \{0,1\},\ 1 \leq j \leq m \end{cases} \qquad (6)$$

According to Eq. (5), the value is calculated in terms of items. Eq. (6) ensures that the total weight of the loaded elements does not exceed the capacity *C* of the knapsack and ensures that each element in the knapsack belongs to an item that is already in the knapsack.

In the scheme where the element is the basic loading unit, the binary vector $\mathbf{Y} = [y_1, y_2, \ldots, y_n]$ of an individual specifies whether or not each element is loaded. It is only a feasible solution if it satisfies the conditions in Eq. (6). Similarly, in element-based loading algorithms, repairing and optimizing solutions are essential in solving SUKP. The following describes the self-adjusting repair and optimization operator that is based on element loading. Each element's value depends on the items to which it belongs because the element itself has no direct value. We assign each element a value according to the proportion of its weight for each item to which it belongs. The absolute value of an element is the sum of its occupation values in the items to which it belongs. The value of each item is only assigned to the elements that are outside the knapsack because we select from only among the elements that are outside the knapsack. It is the most responsive way to set the dynamic values of the elements. Then, the elements that are outside the knapsack have relative value and relative value density, which are defined as follows.

**Define**:

**Absolute value of an element** (*AVE*): The sum of the values of the element in the items to which it belongs.

$$AVE = \sum_{e_j \in \mathbf{U}_i} (p_i / \sum_{e_k \in \mathbf{U}_i} w_k) * w_j \tag{7}$$

**The relative value of an element** (*RVE*): The sum of the values of the element in the items to which it belongs, provided that the value of each item is only assigned to elements that are outside the knapsack.

$$RVE = \sum_{e_j \in U_i} (p_i / \sum_{e_k \in U_i, e_k \notin B} w_k) * w_j \tag{8}$$

**The relative value density of an element** (*RVDE*): The ratio between the *RVE* of element $e_j$ and the weight of element $e_j$

$$RVDE = \sum_{e_j \in U_i} (p_i / \sum_{e_k \in U_i, e_k \notin B} w_k) \tag{9}$$

Where in Eq. 7 to Eq. 9, $w_j$ is the weight of element $e_j$; $p_i$ is the value of item $\mathbf{U}_i$; and $\mathbf{B}$ is the set of elements that are in the knapsack.

ESRO repairs and optimizes the solution according to *RDVE* of the elements. ESRO is described in Algorithm 2. Let **B** be the set of elements that are in the knapsack and let **A** still denote the set of items that have been loaded into the knapsack. Let W(**B**) represent the total weight of the elements that have been loaded into the knapsack. Let **Z** = [$z_1$, $z_2$, ..., $z_m$], $z_j \in \{0, 1\}$, where $1 \leq j \leq m$, be a vector that specifies the loading of the items.

---

**Algorithm 2:** ESRO

**Input:** The candidate solution **Y** = [$y_1$, $y_2$, ..., $y_n$], where $y_j \in \{0, 1\}$ and $1 \leq j \leq n$.

**Output:** The optimized feasible solution **Y** = [$y_1$, $y_2$, ..., $y_n$], where $y_j \in \{0, 1\}$ and $1 \leq j \leq n$, and **Z** = [$z_1$, $z_2$, ..., $z_m$], where $z_j \in \{0, 1\}$ and $1 \leq j \leq m$; the total value of the items in the knapsack, namely, the value of the objective function.

---

1). **B** = Φ, **A** = Φ

2). Let all elements with a value of 1 in **Y** form set **S**1 and let all elements with a value of 0 in **Y** form set **S**2; let **Y** and **Z** both be zero vectors.

3). *while* (**S**1 ≠ Φ)

4).     Identify the element $e_t$ with the highest *RVDE* in **S**1

5).     *if* ($w_t$ + W(**B**) ≤ C)   *then* **B** = **B** ∪ { $e_t$ }

6).     **S**1 = **S**1 - { $e_t$ }

7). *end while*

8). *while* (**S**2 ≠ Φ)

9).     Identify the element $e_t$ with the highest *RVDE* in **S**2

10).    *if* ($w_t$ + W(**B**) ≤ C)   *then* **B** = **B** ∪ { $e_t$ }

11).    **S**2 = **S**2 - { $e_t$ }

12). *end while*

13). *for i* = 1 *to n*

14).    *if* ( **U**$_i$ ⊆ **B** ) *then*  **A** = **A** ∪ {**U**$_i$}

15). *end for*

16). Delete all elements that belong to **B** and do not belong to any item in **A.** If no element is deleted, *goto* 28

>     17). **Q = E - B**
>     18). *while* ( **Q** ≠ Φ )
>     19).     Identify the element $e_t$ with the highest *RVDE* in **Q**
>     20).     *if* $(w_t + W(\mathbf{B}) \leq C)$   *then* **B = B** ∪ { $e_t$ }
>     21).     **Q = Q** - { $e_t$ }
>     22). *end while*
>     23). *flag* = 0
>     24). *for* $\forall \mathbf{U}_i \notin \mathbf{A}$   *do*
>     25).     *if* $(\mathbf{U}_i \subseteq \mathbf{B})$   *then* **A = A** ∪ {$\mathbf{U}_i$},   *flag* =1
>     26). *end for*
>     27). *if* (*flag* = 1) *then goto* 16
>     28). *return* (**Y**, **Z**, P(**A**))

ESRO selects **Y** to become a feasible solution by conducting step 3 to step 16 and optimizes **Y** to obtain a higher quality solution by executing step 17 to step 26. When using the element-based loading method, the solution vector **Y** represents the selected elements and vector **Z** represents the loading of the items. Users can use any element according to their needs.

## 3. The general binary framework for solving SUKP

The above self-adjusting repair optimization operator is specially designed for using swarm intelligence algorithm to solve SUKP. TLBO algorithm has been shown excellent performance in solving many optimization problems. In this work, we use the modified TLBO and the self-adjusting repair optimization operator to solve SUKP. In the following, we propose a general binary TLBO framework (DTLBO), so that both the item-based loading method and the element-based loading method can solve SUKP basing on it. In DTLBO, we introduce elite opposite search and natural elimination mechanisms to improve the search capabilities.

## 3.1 The DTLBO overview

The original TLBO is designed for continuous optimization problems (Rao et al. 2011a; Rao et al. 2011b). In order to solve SUKP effectively, we discretize TLBO and improves it. Algorithm 3 is the framework of the DTLBO algorithm for solving SUKP. Note that the individual in the algorithm is represented by a triplet, please refer to subsect 3.2 for the individual coding.

**Algorithm 3: Discrete TLBO algorithm**
**Input:** SUKP instance, maximal number of function evaluations (*MFC*).
**Output**: The best individual and its fitness.

1). Generate initial real vectors $X_i$, where $i = 1, 2, …, popsize$, randomly, and generate binary vectors $Y_i$, where $i = 1, 2, …, popsize$, that correspond to each real vector via Eq. 10.
2). Repair and optimize all individuals via ISRO or ESRO and calculate their finesses. $X_i$ is changed according to each $Y_i$, $i = 1, 2, …, popsize$, via Eq. 11
3). $t = popsize$.
4). Select the fittest individual as the teacher ($X_{teacher}$, $Y_{teacher}$, $Z_{teacher}$).
5). ***while*** ($t <= MFC$)
6). **Teacher phase:**
7).     For each student individual
8).         Obtain temp real vector $X_{temp}$ via Eq. 12 and temp binary vector $Y_{temp}$ via Eq. 10
9).         Repair and optimize the temp individual and calculate the fitness value of the temp individual.
10).        $X_{temp}$ is changed according to $Y_{temp}$ via Eq. 11. $Z_{temp}$ is obtained according to $Y_{temp}$.
11).        If the temp individual is more fit than the old individual, the old individual is replaced by the temp individual.
12).        $t = t+1$. **if** ($t > MFC$) **go to** 24
13). **Learner phase:**
14).    For each student individual
15).        Obtain temp real vector $X_{temp}$ via Eq. 13 and obtain temp binary vector $Y_{temp}$ via Eq. 10
16).        Repair and optimize the temp individual ($X_{temp}$, $Y_{temp}$, $Z_{temp}$) and calculate the fitness of the temp individual.
17).        $X_{temp}$ is changed according to each $Y_{temp}$ via Eq. 11. $Z_{temp}$ is obtained according to $Y_{temp}$.

18). If the temp individual is more fit than the old individual, the old individual is replaced by the temp individual.
19). $t = t+1$. **if** ($t >MFC$) **go to** 24
20). EOS(). // Elite opposite search, see subsect 3.4
21). $t = t+1$. **if** ($t >MFC$) **go to** 24
22). SF(). //Survival of the fittest, see subsect 3.4
23). $t = t+1$. **if** ($t >MFC$) **go to** 24
24). Select the fittest individual as the teacher ($X_{teacher}$, $Y_{teacher}$, $Z_{teacher}$).
25). *end while*
26). *return* ($X_{teacher}$, $Y_{teacher}$, $Z_{teacher}$, $f(Y_{teacher})$).

The $f(Y_{teacher})$ in step 26 is the fitness value of the fittest individual. The Algorithm 3 is a generic framework for solving SUKP and it is available for any of the loading methods that are proposed in Section 2. If items are used as the loading units, the individual triad (**X**, **Y**, **Z**) is defined as in Figure 2(b). If elements are used as the loading units, the individual triad (**X**, **Y**, **Z**) is defined as in Figure 2(c).

### 3.2 Encoding Schema

In this work, we use the encoding transformation method (He et al. 2019) to design a general discrete TLBO algorithm for two loading methods. This requires that the encoding schema must meet the requirements of the two loading methods. Let each individual in TLBO be represented by a triad (**X**, **Y**, **Z**). $\mathbf{X} = [x_1, x_2, …, x_d]$, where $x_j \in [0, 1]$, with $1 \leq j \leq d$, is a real vector, and $\mathbf{Y} = [y_1, y_2, …, y_d]$, $y_j \in \{0, 1\}$, where $1 \leq j \leq d$, and $\mathbf{Z} = [z_1, z_2, …, z_g]$, where $z_k \in \{0, 1\}$ and $1 \leq k \leq g$, are both binary vectors. The real vector **X** and binary vector **Y** are mapped to each other according to Eq. 10 and Eq. 11. The evolution of the algorithm is conducted on the real vector **X**. Binary vector **Y** is the solution of SUKP and the fitness of each individual is calculated according to it. In the approach of loading in units of items, **Y** represents the loading of the items and Z represents element vector that corresponds to **Y**. In the loading method in units of elements, **Y** represents the loading of the elements and **Z** is the item vector that corresponds to **Y**.

$$y_j = \begin{cases} 1 & x_j > 0 \\ 0 & others \end{cases} \quad j = 1, 2, ..., d \tag{10}$$

$$x_j = \begin{cases} rand(0, 1) & y_j = 1 \\ rand(0, 1) - 1 & others \end{cases} \quad j = 1, 2, ..., d \tag{11}$$

The following example illustrates the relationship among **X**, **Y** and **Z** in each individual in the discrete TLBO. Consider the following SUKP instance: **E** = {$e_1$, $e_2$, ..., $e_6$} and **S** = {$U_1$, $U_2$, $U_3$, $U_4$}. **S** is expressed in Figure 2(a). $S_{ij}$= 1 in the matrix if $e_j \in U_i$; $S_{ij}$= 0 in the matrix if $e_j \notin U_i$. Figure 2(b) illustrates the correspondence among the three vectors of an individual when the method of loading items is adopted. Figure 2(c) illustrates the correspondence among the three vectors in an individual when the method of loading elements is adopted. In vectors **Y** and **Z**, 1 denotes that the corresponding object is a loading object and 0 indicates that it is not a loading object. In Figure 2(b), **Y** represents the loading of the item; its value is determined by **X** and its value determines the value of **Z**. In Figure 2(c), **Y** represents the loading of the elements; its value is determined by **X** and its value determines the value of **Z**.

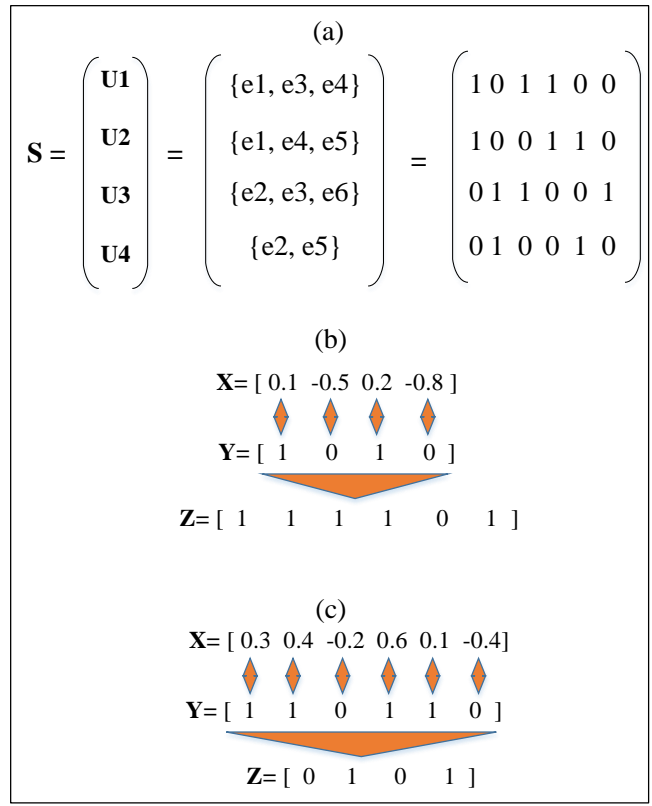

**Fig. 2** Example relation among the three vectors of an individual in Discrete TLBO. (a) The considered instance (b) Items as loading units (c) Elements as loading units

## 3.3 Teacher phase and Learner phase

The entire algorithm of TLBO is divided into Teacher phase and Learner phase (Rao et al. 2011a; Rao et al. 2011b). The individuals in the population move toward the optimal individual in the Teacher phase, and the exploration of the algorithm is realized by the students learning from each other in Learner phase. The following is the evolution process of two phases, which the individual is only a real vector. In the DTLBO, the real vectors of individuals still evolve according to these formulas.

In the Teacher phase, each student learns according to the gap between the average grade and the teacher, which is expressed in Eq. (12) (Rao et al. 2011a; Rao et al. 2011b). If the individual after studying has better fitness, the new individual is accepted.

$$\mathbf{X}_i' = \mathbf{X}_i + rand(0,1)*(\mathbf{X}_{teacher} - T_F \mathbf{X}_{mean}) \tag{12}$$

In the Learner phase, the students learn from each other according to Eq. (13) (Rao et al. 2011a; Rao et al. 2011b). If an individual after learning has better fitness, the new individual is accepted; otherwise, the individual is unchanged.

$$\mathbf{X}_i' = \begin{cases} \mathbf{X}_i + rand(0,1)*(\mathbf{X}_i - \mathbf{X}_k) & if\ fit(\mathbf{X}_i) > fit(\mathbf{X}_k) \\ \mathbf{X}_i + rand(0,1)*(\mathbf{X}_k - \mathbf{X}_i) & others \end{cases} \tag{13}$$

## 3.4 New search strategy

In real life, teachers usually continue to learn for improving themselves. When their works enter a certain bottleneck, opposite thinking is a common way they use to break out of their limitations. Inspired by this, we let elite individual (the teacher) in DTLBO implement an opposite search in each iteration. This greatly improves the exploration capabilities of swarm population. We call the search as function EOS( ) (Elite opposite search). The EOS() process is as follow:

1). Get a new temporary individual *temp* by inverting each bit of the teacher's binary vector, that is, 0 becomes 1, and 1 becomes 0.

2). Use repair and optimization operator to handle *temp*.

3). If *temp* is better than teacher, *temp* replace teacher, otherwise do nothing.

The diversity of population can help the algorithm search the search space more comprehensively, thereby improving the accuracy of the solution. In order to increase the diversity of the population, we introduce a natural elimination mechanism. It is that each iteration will eliminate the worst individual in the population, and replace it with a randomly generated new individual. The addition of new random individuals can effectively avoid premature of algorithm. This process is represented by the function SF( ) (Survival of the fittest). This seemingly simple strategy can continuously inject new life into the population and enhance the global search ability.

## 4. Simulation and evaluation

### 4.1 Experimental description

The 30 benchmark instances that are considered in the current studies (Feng et al. 2018; He et al. 2018a; He et al. 2018b; Liu and He 2019; Ozsoydan and Adil 2018; Wei and Hao 2019; Wu and He 2020) of SUKP were generated in reference (He et al. 2018a). We evaluate the performances of the proposed operators and algorithm on these 30 instances. These instances are named sukp m_n_α_β according to the generated rule, where $m$ is the number of items, $n$ is the number of elements, α is the density of the elements, and β is the ratio of the capacity to the total weight of the elements. There are three types of SUKP instances according to the relationship between $m$ and $n$ ($m>n$, $m=n$, $m<n$). These benchmark instances can be obtained at http://sncet.com/ThreetypesofSUKPinstances(EAs).rar.

In previous research (Feng et al. 2018; He et al. 2018a; He et al. 2018b; Liu and He 2019; Ozsoydan and Adil 2018; Wei and Hao 2019; Wu and He 2020), the maximum number of iterations $Maxiter=max\{m, n\}$ ($m$ is the number of items and $n$ is the number of elements) is used as the termination condition of the algorithm, and the size of the population is set as 20. For fair comparison, we terminate the algorithm when the maximal number of function evaluations, namely, $MFC=20+20* max \{m, n\}$, is reached. All experiments are conducted in Visual C++ 2010 on a PC that is equipped with an

Intel(R) Core(TM) i5-7Y54 CPU@1.2 GHz with 8 G of RAM in the Windows 10 operating system.

## 4.2 Comparison and analysis of experimental results

Two self-adjusting repair and optimization operators are proposed in this paper, which are applicable to the two loading methods. We use the DTLBO algorithm with two new search strategies to solve SUKP. If the item is the basic unit for loading, the repair and optimization operator use ISRO. We refer to the algorithm as I-DTLBO. If the element is the basic unit for loading, the repair and optimization strategy use ESRO. We refer to the algorithm as E-DTLBO. We compare I-DTLBO and E-DTLBO with the experimental results of the current excellent swarm intelligence algorithms (BABC (He et al. 2018a; He et al. 2018b), GROA (He et al. 2018b), gPSO (Ozsoydan and Adil 2018) and DHJaya (Wu and He 2020)) for solving SUKP. The experimental results of all algorithms are shown in Tables 1-3.

### 4.2.1 Computational Results and Comparisons

In Tables 1-3, we compare and analyze the experimental data according to the best value (Best), the worst value (Worst), the average value (Mean) and the standard deviation (Std). The bold data in these tables indicate the best solution among several algorithms.

According to Table 1 to Table 3, I-DTLBO performed very well. I-DTLBO reached the upper boundary of the current swarm intelligence algorithms for solving SUKP in 10 instances, and gotten new upper boundary in 15 instances. For 27 instances, I-DTLBO gotten the best in Mean among these algorithms. In the worst value comparison, I-DTLBO is the best in 29 instances. These data fully illustrate the excellence of the I-DTLBO. And in sharp contrast is the E-DTLBO algorithm based on the element as the basic loading unit. It can get well solutions in small-scale and middle-scale instances, but it performs very badly for large-scale instances. The performance of E-DTLBO completely negated our original intuitive guess. It shows that the loading strategy using elements is not suitable for solving SUKP, although the element is the smallest unit that carries weight.

Tab. 1  The computing results of the first kind of SUKP instances (The experimental results of BABC are obtained from He et al.2018a and He et al.2018b, data of gPSO are from Ozsoydan et al.2018, data of DHJava are from Wu and He 2020, data of GTOA are from He et al.2018b)

| Instance | Result | BABC | gPSO | DHJaya | GTOA | I-DTLBO | E-DTLBO |
|---|---|---|---|---|---|---|---|
| sukp100_85_0.10_0.75 | Best | 13251 | **13283** | **13283** | 13251 | **13283** | 13251 |
|  | Mean | 13028.5 | 13050.53 | 13066.3 | 13025.4 | **13237.6** | 12949.3 |
|  | Worst | --------- | **13044** | 13016 | 12763 | 12918 | 12622 |
|  | Std | 92.63 | 37.41 | 56.24 | 92.63 | 84.96 | 130.05 |
| sukp100_85_0.15_0.85 | Best | 12238 | **12774** | 12348 | 12274 | 12479 | 12479 |
|  | Mean | 12155 | 12084.82 | 12156.4 | 12029.3 | 12277.6 | **12370.7** |
|  | Worst | --------- | 11631 | 12065 | 11483 | **12100** | 12065 |
|  | Std | 53.29 | 95.38 | 66.35 | 127.25 | 126.43 | 133.25 |
| Sukp200_185_0.10_0.75 | Best | 13241 | 13405 | 13405 | 13405 | **13521** | 13502 |
|  | Mean | 13064.4 | 13286.56 | 13254.8 | 13196.9 | **13405.6** | 13165.3 |
|  | Worst | --------- | 12862 | 12948 | 12932 | **13244** | 12819 |
|  | Std | 99.57 | 93.18 | 101.57 | 127.43 | 72.22 | 168.23 |
| Sukp200_185_0.15_0.85 | Best | 13829 | 14044 | **14215** | **14215** | **14215** | 13993 |
|  | Mean | 13359.2 | 13492.60 | 13534.7 | 13285.9 | **14018.4** | 13397.1 |
|  | Worst | --------- | 12785 | 13022 | 12763 | **13706** | 12660 |
|  | Std | 234.99 | 328.72 | 287.65 | 276.35 | 173.41 | 315.16 |
| Sukp300_285_0.10_0.75 | Best | 10428 | 11335 | 10869 | 11407 | **11563** | 10966 |
|  | Mean | 9994.7 | 10669.51 | 10683.5 | 10674.5 | **11396.6** | 10444.4 |
|  | Worst | --------- | 9900 | 10225 | 10032 | **11047** | 9904 |
|  | Std | 154.03 | 227.85 | 137.95 | 599.55 | 131.14 | 255.37 |
| Sukp300_285_0.15_0.85 | Best | 12012 | 12245 | 12245 | 12245 | **12401** | 12044 |
|  | Mean | 10902.9 | 11607.10 | 11801.4 | 11533.4 | **12058.4** | 11491.4 |
|  | Worst | --------- | 10536 | 10940 | 10382 | **11506** | 10955 |
|  | Std | 449.45 | 477.80 | 370.29 | 466.27 | 173.47 | 238.19 |
| Sukp400_385_0.10_0.75 | Best | 10766 | **11484** | 11321 | 11435 | **11484** | 10498 |
|  | Mean | 10065.2 | 10915.87 | 10866.1 | 10830.3 | **11135.3** | 9971.1 |
|  | Worst | --------- | 10158 | 10086 | 10051 | **10753** | 9565 |
|  | Std | 241.45 | 367.75 | 332.88 | 353.33 | 178.32 | 225.53 |
| Sukp400_385_0.15_0.85 | Best | 9649 | 10710 | 10175 | 10397 | **11029** | 10642 |
|  | Mean | 9135.98 | 9864.55 | 9708.2 | 9894.6 | **10521.5** | 10147.9 |
|  | Worst | --------- | 9173 | 9253 | 9142 | **9996** | 9710 |
|  | Std | 151.90 | 315.38 | 242.91 | 290.02 | 289.65 | 197.85 |
| Sukp500_485_0.10_0.75 | Best | 10784 | 11722 | 11546 | 11716 | **11771** | 10163 |
|  | Mean | 10452.2 | 11184.51 | 11071.8 | 11171.4 | **11657.2** | 9482.2 |
|  | Worst | --------- | 10614 | 10553 | 10591 | **11435** | 9126 |
|  | Std | 114.35 | 322.98 | 254.33 | 299 | 67.64 | 205.80 |
| Sukp500_485_0.15_0.85 | Best | 9090 | 10022 | 9625 | 9860 | **10194** | 8783 |
|  | Mean | 8857.89 | 9299.56 | 9198.4 | 9262 | **9840.26** | 8415.1 |
|  | Worst | --------- | 8556 | 8893 | 8742 | **9625** | 7853 |
|  | Std | 94.55 | 277.62 | 167.73 | 153.89 | 112.38 | 213.87 |

Tab. 2    The computing results of the second kind of SUKP instances (The experimental results of BABC are obtained from He et al.2018a and He et al.2018b, data of gPSO are from Ozsoydan et al.2018, data of DHJava are from Wu and He 2020, data of GTOA are from He et al.2018b)

| Instance | Result | BABC | gPSO | DHJaya | GTOA | I-DTLBO | E-DTLBO |
|---|---|---|---|---|---|---|---|
| sukp100_100_0.10_0.75 | Best | 13860 | **14044** | **14044** | **14044** | **14044** | 13957 |
|  | Mean | 13734.9 | 13854.71 | 13876.5 | 13792.5 | **13989.1** | 13620.6 |
|  | Worst | --------- | 13664 | 13668 | 13561 | **13889** | 13199 |
|  | Std | 70.76 | 96.23 | 68.16 | 90.57 | 33.70 | 169.87 |
| sukp100_100_0.15_0.85 | Best | **13508** | **13508** | **13508** | **13508** | **13508** | **13508** |
|  | Mean | 13352.4 | 13347.58 | 13435.7 | 13220.5 | **13502.4** | 13472.6 |
|  | Worst | --------- | 12613 | 13280 | 11988 | **13407** | 13161 |
|  | Std | 155.14 | 194.34 | 54.73 | 296.61 | 19.80 | 76.82 |
| Sukp200_200_0.10_0.75 | Best | 11846 | **12522** | **12522** | 12350 | 12384 | 12089 |
|  | Mean | 11194.3 | 11898.73 | 11921.2 | 11983.4 | **12128.9** | 11720.1 |
|  | Worst | --------- | 11048 | 11050 | 10903 | **11853** | 11423 |
|  | Std | 249.58 | 391.83 | 325.85 | 326.08 | 123.16 | 176.75 |
| Sukp200_200_0.15_0.85 | Best | 11521 | **12317** | 11846 | **12317** | **12317** | 12191 |
|  | Mean | 10945 | 11584.64 | 11632.6 | 11572.8 | **11936.7** | 11687.4 |
|  | Worst | --------- | 10510 | 11291 | 10957 | **11624** | 11276 |
|  | Std | 255.14 | 391.83 | 190.27 | 242.36 | 142.70 | 213.84 |
| Sukp300_300_0.10_0.75 | Best | 12186 | 12695 | 12646 | 12695 | **12715** | 11814 |
|  | Mean | 11945.8 | 12411.27 | 12329.1 | 12464.1 | **12559**.9 | 11427.8 |
|  | Worst | --------- | 10510 | 11871 | 11968 | **12231** | 11094 |
|  | Std | 127.80 | 275.32 | 237.57 | 231.68 | 133.58 | 175.61 |
| Sukp300_300_0.15_0.85 | Best | 10382 | 11425 | 10884 | 11425 | **11585** | 11191 |
|  | Mean | 9859.69 | 10568.41 | 10575 | 10513.9 | **11192.3** | 10937 |
|  | Worst | --------- | 9648 | 9936 | 9477 | **10888** | 10132 |
|  | Std | 177.02 | 327.48 | 175.31 | 355.88 | 146.71 | 192.30 |
| Sukp400_400_0.10_0.75 | Best | 10626 | 11531 | 11128 | 11450 | **11665** | 10841 |
|  | Mean | 10101.1 | 10958.96 | 10722.2 | 10951.9 | **11352.9** | 10086.9 |
|  | Worst | --------- | 10205 | 10305 | 10360 | **11029** | 9183 |
|  | Std | 196.99 | 274.90 | 208.18 | 264.34 | 150.92 | 442.60 |
| Sukp400_400_0.15_0.85 | Best | 9541 | 10927 | 10915 | 10915 | **11325** | 9994 |
|  | Mean | 9032.95 | 9845.17 | 9711.6 | 9834.4 | **10595.9** | 8961.8 |
|  | Worst | --------- | 9033 | 9258 | 9145 | **10060** | 8486 |
|  | Std | 194.18 | 358.91 | 265.33 | 312.87 | 374.50 | 317.26 |
| Sukp500_500_0.10_0.75 | Best | 10755 | 10888 | 10835 | 10960 | **11249** | 9407 |
|  | Mean | 10328.5 | 10681.46 | 10604.6 | 10626.1 | **10857.9** | 8862 |
|  | Worst | --------- | 10222 | 10313 | 10048 | **10647** | 8484 |
|  | Std | 91.615 | 125.36 | 112.87 | 152.96 | 101.72 | 227.89 |
| Sukp500_500_0.15_0.85 | Best | 9318 | 10194 | 10176 | 10194 | **10202** | 9132 |
|  | Mean | 9180.74 | 9703.62 | 9706.7 | 9754.7 | **10006.8** | 8387 |
|  | Worst | --------- | 8892 | 9348 | 9044 | **9780** | 7864 |
|  | Std | 84.91 | 252.84 | 216.28 | 231.77 | 117.98 | 289.01 |

Tab. 3 The computing results of the third kind of SUKP instances (The experimental results of BABC are obtained from He et al.2018a and He et al.2018b, data of gPSO are from Ozsoydan et al.2018, data of DHJava are from Wu and He 2020, data of GTOA are from He et al.2018b)

| Instance | Result | BABC | gPSO | DHJaya | GTOA | I-TLBO | E-TLBO |
|---|---|---|---|---|---|---|---|
| sukp85_100_0.10_0.75 | Best | 11664 | **12045** | 11851 | **12045** | **12045** | **12045** |
| | Mean | 11182.7 | 11486.95 | 11442.5 | 11388.3 | 11790.8 | **11903.5** |
| | Worst | --------- | 11202 | 11287 | 11083 | **11619** | 11522 |
| | Std | 183.57 | 137.52 | 97.78 | 107.48 | 102.53 | 123.77 |
| sukp85_100_0.15_0.85 | Best | **12369** | 12369 | 12369 | 12369 | 12369 | 12369 |
| | Mean | 12081.6 | 11994.36 | 12246.1 | 11945.7 | 12318.7 | **12320.9** |
| | Worst | --------- | 11274 | 11413 | 11251 | **12124** | **12124** |
| | Std | 193.79 | 436.81 | 191.31 | 400.84 | 79.94 | 74.63 |
| Sukp185_200_0.10_0.75 | Best | 13047 | **13696** | 13647 | 13647 | **13696** | 13529 |
| | Mean | 12522.8 | 13204.26 | 13181.2 | 13143.5 | **13389** | 13114.2 |
| | Worst | --------- | 12339 | 12584 | 12170 | **13001** | 12708 |
| | Std | 201.35 | 436.81 | 231.47 | 308.38 | 154.15 | 178.43 |
| Sukp185_200_0.15_0.85 | Best | 10602 | **11298** | 11083 | 10973 | **11298** | 10946 |
| | Mean | 10150.6 | 10801.41 | 10711.1 | 10566.1 | **11022.5** | 10661.3 |
| | Worst | --------- | 10195 | 10142 | 9790 | **10688** | 10293 |
| | Std | 152.91 | 205.76 | 183.62 | 292.99 | 186.03 | 173.74 |
| Sukp285_300_0.10_0.75 | Best | 11158 | **11568** | 11568 | 11568 | 11538 | **11568** |
| | Mean | 10775.9 | 11317.99 | 11241.9 | 11202.4 | **11296** | 10420 |
| | Worst | --------- | 10690 | 10854 | 10734 | **11061** | 9959 |
| | Std | 116.80 | 182.82 | 141.91 | 201.21 | 110.82 | 248.75 |
| Sukp285_300_0.15_0.85 | Best | 10528 | 11517 | 11237 | 11377 | 11763 | **11802** |
| | Mean | 9897.92 | 10899.20 | 10649.9 | 10821.9 | **11409.1** | 11202.1 |
| | Worst | --------- | 10281 | 10083 | 10004 | **11006** | 10362 |
| | Std | 186.53 | 300.36 | 276.81 | 319.23 | 183.25 | 422.28 |
| Sukp385_400_0.10_0.75 | Best | 10085 | **10483** | 10157 | 10326 | 10332 | 9355 |
| | Mean | 9537.5 | 10013.43 | 9878.7 | 9949.3 | **10155.8** | 8797.0 |
| | Worst | --------- | 9519 | 9474 | 9501 | **9968** | 8417 |
| | Std | 184.62 | 202.40 | 150.75 | 165.8 | 81.41 | 200.54 |
| Sukp385_400_0.15_0.85 | Best | 9456 | 10338 | 10075 | 10302 | **10359** | 10338 |
| | Mean | 9090.03 | 9524.98 | 9443.7 | 9381.3 | **10059.1** | 9070.9 |
| | Worst | --------- | 8816 | 9016 | 8841 | **9634** | 8381 |
| | Std | 156.69 | 286.16 | 242.74 | 286.24 | 138.28 | 456.75 |
| Sukp485_500_0.10_0.75 | Best | 10823 | 11094 | 10913 | 11037 | **11285** | 10143 |
| | Mean | 10483.4 | 10687.62 | 10709.3 | 10658 | **10907.4** | 9099 |
| | Worst | --------- | 10201 | 10408 | 10206 | **10646** | 8570 |
| | Std | 228.34 | 168.06 | 123.19 | 164.93 | 110.47 | 318.12 |
| Sukp485_500_0.15_0.85 | Best | 9333 | 10104 | 9642 | 9964 | **10120** | 8802 |
| | Mean | 9085.57 | 9383.28 | 9383.4 | 9356 | **9817.5** | 8174.0 |
| | Worst | --------- | 8834 | 9030 | 8785 | **9482** | 7727 |
| | Std | 115.62 | 241.01 | 124.69 | 205.87 | 142.49 | 252.68 |

For swarm intelligence algorithms, the mean is more representative of the algorithm's solving performance. We use Friedman Test and Nemenyi Test to test the experimental mean results to more clearly prove the I-DTLBO to be effective. The Friedman Test is a rank-based test that assumes that the ranks of all samples are equal. Specifically, we first sort the mean values of different algorithms on 30 instances, and finally calculate the average of the ranking of each algorithm on all the instances. The average ranks of six algorithms over all instances are shown in Table 4. The *p-values* < 0.05 indicates that the null hypothesis is rejected and there are significant differences between the algorithms.

Tab. 4    Friedman Test ranks on mean and corresponding *p-values*

| Algorithms | Average ranks |
|---|---|
| **I-DTLBO** | 1.133 |
| **E-DTLBO** | 4.533 |
| **BABC** | 5.233 |
| **gPSO** | 3.0 |
| **GTOA** | 3.966 |
| **DHJaya** | 3.2 |
| *p-value* | 0.000 |

Because the Friedman test can only show that there are significant differences among these algorithms, but cannot determine which algorithms have differences, we further use the Nemenyi test for post-hoc analysis. We calculated the critical difference (*CD*) for the average ordinal value difference as 1.376 at the significance level $\alpha = 0.05$. Table 5 is the calculated ordinal difference between I-DTLBO and other algorithms. These results are compared with *CD*. When the value is greater than *CD*, it indicates that there is a significant difference between the two algorithms, otherwise there is no difference. Table 5 shows the significant difference between I-DTLBO and other algorithms. It can be concluded from Tables 4 and 5 that I-DTLBO is significantly better than other algorithms.

Tab. 5  Ordinal difference between I-TLBO and other algorithms

| Algorithms | Ordinal difference | Comparison | Significant difference |
|---|---|---|---|
| E-DTLBO | 3.4 | >CD | Yes |
| BABC | 4.100 | > CD | Yes |
| gPSO | 1.866 | > CD | Yes |
| GTOA | 2.833 | > CD | Yes |
| DHJaya | 2.066 | >CD | Yes |

**4.2.2 Effectiveness of self-adjusting repair and optimization operator**

In order to test the performance of the two repairing and optimization operators, we compared these two operators with the existing two operators under the same conditions. We combine DTLBO with ISOR, ESOR, S-GROA (He et al. 2018a) and SM-GROA (Wu and He 2020) respectively to solve SUKP. They are named I-TLBO, E-TLBO, S-TLBO and MS-TLBO. The results for I-TLOB, E-TLBO, S-TLBO and MS-TLBO are obtained from 50 independent runs. We use the boxplot of statistics to analyze the experimental results of them to compare clearly the performance of the four repair and optimization strategies. As we can see from Figure 3-5, it is clear that the overall boxplot of I-TLBO is the highest, which indicate that the accuracy of I-TLBO is the best; the much data above the median, and the small interquartile range and few abnormal data indicate that the stability of the algorithm is very great. ISRO uses relative value density of an item (*RVDI*), and dynamically adjusts the *RVDI* of the remaining items according to the loading items during the loading process, so that the algorithm can accurately select the item that contributes the most to the solution. The E-TLBO algorithm using element-based repair optimization strategy ESRO is superior to S-TLBO and MS-TLBO in solving small-scale problems, but the searching ability in large-scale instances is obviously lower than other algorithms. Element-based loading can reflect the advantages of loading based on the smallest unit when solving small- and medium-scale instances. However, when the amount of data increases to a certain extent, the information carried by a single element cannot reflect the contribution of element to the solution properly, resulting the decreasing of algorithm's ability.

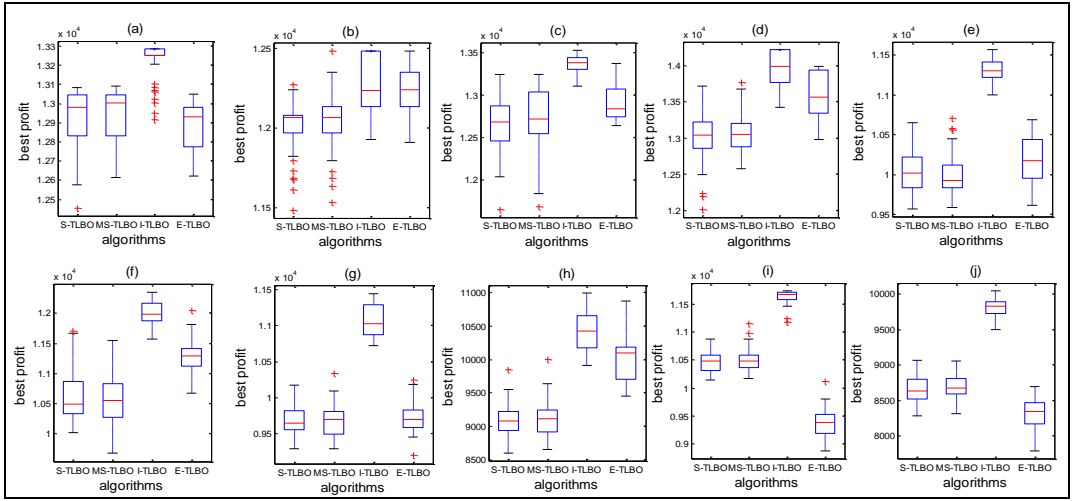

**Fig. 3** Boxplot of S-TLBO, MS-TLBO, I-TLBO and E-TLBO on 10 instances in first kind.

(a) sukp100_85_0.10_0.75 (b) sukp100_85_0.15_0.85 (c) sukp200_185_0.10_0.75

(d) sukp200_185_0.15_0.85 (e) sukp300_285_0.10_0.75 (f) sukp300_285_0.15_0.85

(g) sukp400_385_0.15_0.85 (h) sukp400_385_0.10_0.75 (i) sukp500_485_0.10_0.75

(j) sukp500_485_0.15_0.85

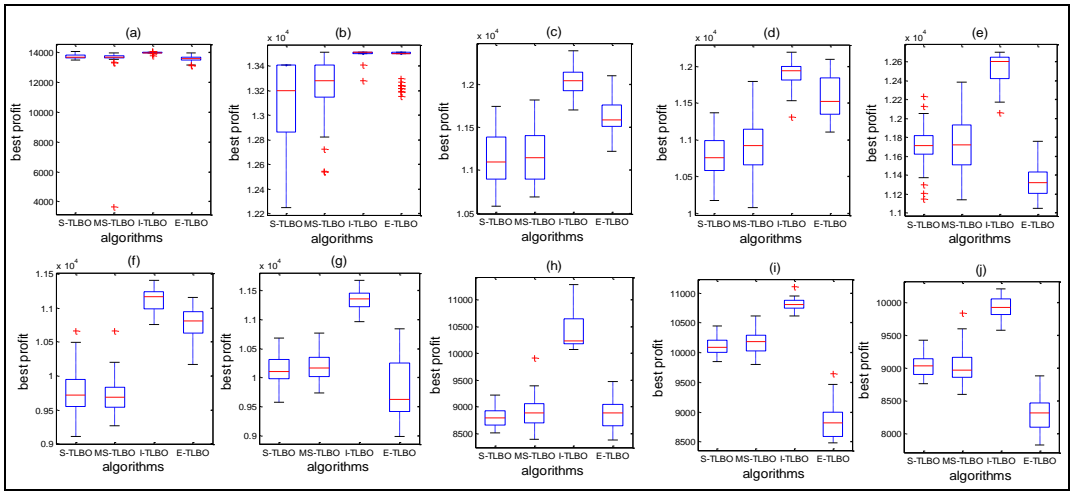

**Fig. 4** Boxplot of S-TLBO, MS-TLBO, I-TLBO and E-TLBO on 10 instances in second kind.

(a) sukp100_100_0.10_0.75 (b) sukp100_100_0.15_0.85 (c) sukp200_200_0.10_0.75

(d) sukp200_200_0.15_0.85 (e) sukp300_300_0.10_0.75 (f) sukp300_300_0.15_0.85

(g) sukp400_400_0.15_0.85 (h) sukp400_400_0.10_0.75 (i) sukp500_500_0.10_0.75

(j) sukp500_500_0.15_0.85

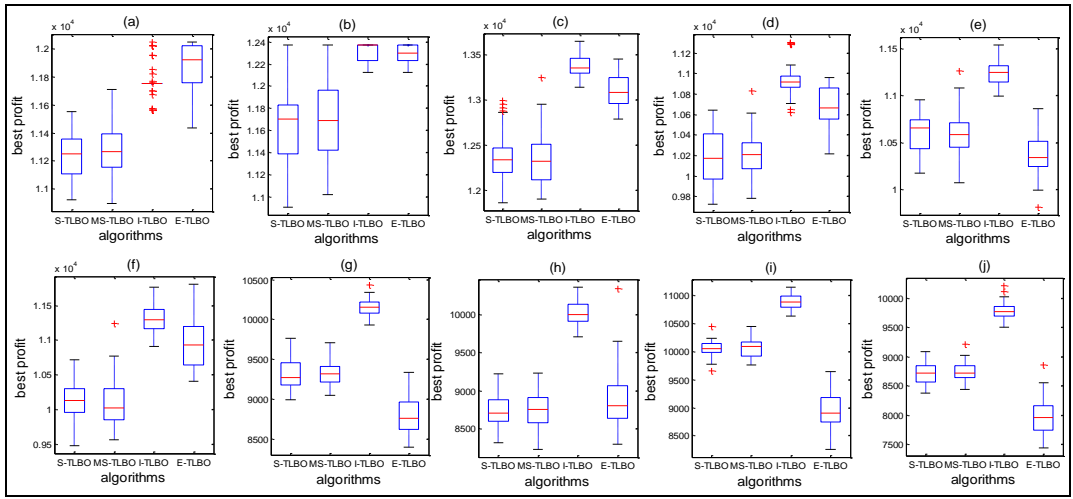

**Fig. 5** Boxplot of S-TLBO, MS-TLBO, I-TLBO and E-TLBO on 10 instances in third kind.

(a) sukp85_100_0.10_0.75 (b) sukp85_100_0.15_0.85 (c) sukp185_200_0.10_0.75

(d) sukp185_200_0.15_0.85 (e) sukp285_300_0.10_0.75 (f) sukp285_300_0.15_0.85

(g) sukp385_400_0.15_0.85 (h) sukp385_400_0.10_0.75 (i) sukp485_500_0.10_0.75

(j) sukp485_500_0.15_0.85

The ISRO outperforms the S-GROA (He et al. 2018a) and MS-GROA (Wu and He 2020) in repairing and optimizing solutions. Because it considers the dynamic influence of the value density of the items on the item loading process, then it uses the method of dynamically adjusting the value density to greedily select the items to be loaded. This method plays a very strong role in improving the accuracy of the solution. Therefore, I-TLBO outperforms the previous algorithms in terms of accuracy and stability. However, as ISRO continuously adjusts the value density of the items, the calculation burden increases.

The ESRO is based on a new loading mode. This is the first study to use elements as loading units. When using this method, the individual in the swarm intelligence algorithm represents the element loading information. The element is the smallest structural unit in SUKP, and we thought it would be more efficient to use the element as a loading unit. However, the experimental results demonstrate that it is not as good as we expected. The elements in SUKP have no direct value, and reasonably allocating the value of the item to its elements is difficult. This is the main reason why E-TLBO perform poorly when solving SUKP.

### 4.2.3 Effectiveness of the new search strategies

In order to improve the searching ability, we add elite opposite search and natural elimination mechanism into DTLBO. The experiment designed below is to verify the effectiveness of the two new search strategies. Except for new search strategies, the other aspects of the two comparisons algorithms are exactly the same. I-DTLBO is the algorithm joins the new search strategy. I-TLBO doesn't use the new search strategy. We compare I-TLBO and I-DTLBO by the accuracy and stability.

The accuracy of solution and stability are two important criteria that reflect the performance of a swarm intelligence algorithm (He et al. 2018b). The mean values in the experimental data show the average performance of the algorithm, and the standard deviation reflects the stability of the algorithm. Figure 6 compares the mean value and standard deviation of experiment results of I-TLBO and I-DTLBO with a two-axis diagram. In Figure 6, the curves are the gap between mean value and the current best (Wei and Hao 2019), they correspond to the right coordinate (mean-gap). The closer the curve is to the X axis, the smaller the gap, and the smaller gap indicates the higher accuracy of the algorithm. The histograms represent the standard deviation, corresponding to the left ordinate (std). The lower the histogram, the smaller standard deviation, and smaller the standard deviation indicates stronger the stability of the algorithm. It is obvious from the three sub-graphs of Figure 6 that I-DTLBO is better than I-TBLO both in accuracy and stability. The two new strategies have achieved the expected goals well.

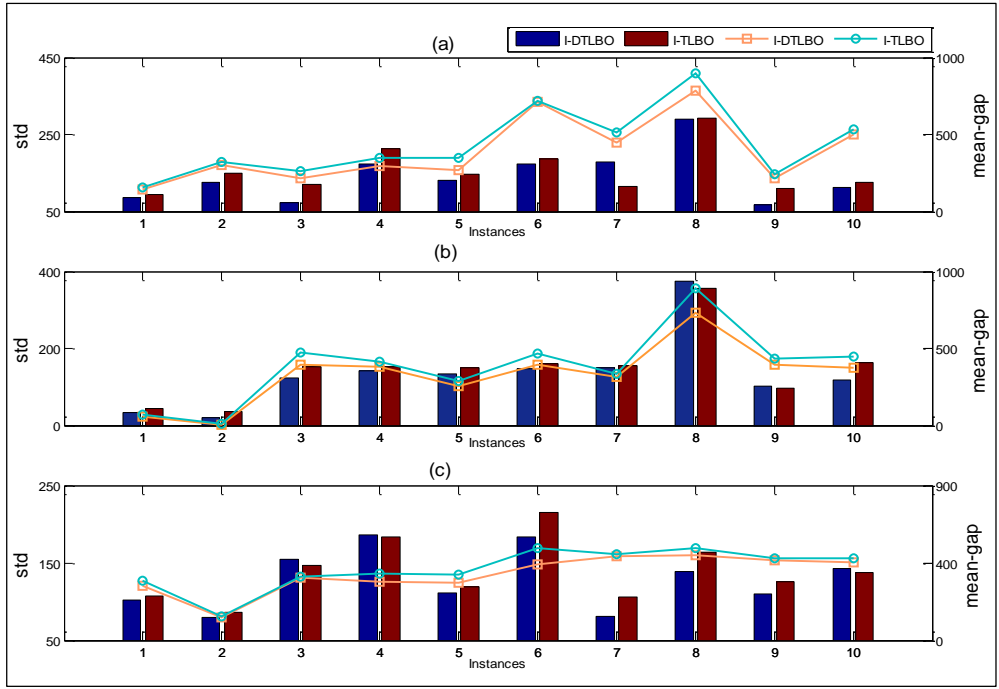

**Fig.6** Comparison of standard deviation and mean-gap between algorithm with new strategy (I-DTLBO) and algorithm without new strategy (I-TLBO): (a) the first kind of SUKP instances (b) the second kind of SUKP instance (c) the third kind of SUKP instances

## 5. Conclusions

Using swarm intelligence algorithm is currently the most popular method to solve SUKP. This paper proposes two algorithms (I-DTLBO and E-DTLBO) based on self-adjusting repair and optimization operators (ISRO and ESRO) to solve SUKP. They are designed from two loading perspectives, item and element, respectively. We propose self-adjusting repair and optimization operator by analyzing the dynamic characteristics of SUKP, which provides a better repair operator for solving SUKP when using swarm intelligence algorithm. In order to improve the searching ability of TLBO, we design elite opposite searching strategy and natural elimination mechanism. The experimental results demonstrate that I-DTLBO exhibits outstanding performance. I-DTLBO using the item-based self-adjusting repair and optimization operator (ISRO) has higher accuracy and stability than other swarm intelligence algorithms. The element-based method we proposed, E-DTLBO, performed well in small and medium-scale instances, and the new upper limit was obtained in 3 instances. But it is obviously inferior to other algorithms in solving large-scale examples. Thinking about solving SUKP from the perspective of elements did not get the expected effect. Compared with the item-based method, the element-based loading method is proved not to apply for solving SUKP.

The self-adjusting repair and optimization operator based on SUKP's dynamic feature is a contribution to solve SUKP using swarm intelligence algorithm, and it can greatly improve the accuracy of the solution. DTLBO with elite opposite searching strategy and natural elimination mechanism is also an effective way to solve other combinatorial optimization problems. There are two plans in the future. One is to use other swarm intelligence algorithms combined with ISRO to solve SUKP. One is to try to use the improved DTLBO to solve other knapsack problems.